\documentclass[11pt,a4paper]{article}
\usepackage[hyperref]{acl2019}
\usepackage{times}
\usepackage{latexsym}

\usepackage{mathrsfs} 
\usepackage{amsmath} 
\usepackage{amsfonts} 

\usepackage{enumerate}

\usepackage{booktabs}
\usepackage{multirow}
\usepackage{floatrow}

\newfloatcommand{capbtabbox}{table}[][\FBwidth]

\usepackage{algorithmic}
\usepackage{algorithm}

\usepackage{graphicx}
\usepackage{subfigure}

\aclfinalcopy 
\def\aclpaperid{1839} 

\title{Open Domain Event Extraction Using Neural Latent Variable Models}

\author{
	Xiao Liu\textsuperscript{1,2} \and Heyan Huang\textsuperscript{1,2} \and Yue Zhang\textsuperscript{3,4}\thanks{\ \ Corresponding author.} \\
	\textsuperscript{1}School of Computer Science and Technology, Beijing Institute of Technology \\
	\textsuperscript{2}Zhejiang Lab, China \\
	{\tt \{xiaoliu,hhy63\}@bit.edu.cn} \\
	\textsuperscript{3}School of Engineering, Westlake University \\
	\textsuperscript{4}Institute of Advanced Technology, Westlake Institute for Advanced Study \\
	{\tt yue.zhang@wias.org.cn}
}

\date{}

\begin{document}
\maketitle
\begin{abstract}

We consider open domain event extraction, the task of extracting unconstraint types of events from news clusters.
A novel latent variable neural model is constructed, which is scalable to very large corpus.
A dataset is collected and manually annotated, with task-specific evaluation metrics being designed.
Results show that the proposed unsupervised model gives better performance compared to the state-of-the-art method for event schema induction.

\end{abstract}

\section{Introduction}

Extracting events from news text has received much research attention.
The task typically consists of two subtasks, namely {\it schema induction}, which is to extract event templates that specify argument slots for given event types \cite{DBLP:conf/emnlp/Chambers13,DBLP:conf/naacl/CheungPV13,DBLP:conf/acl/NguyenTFB15,DBLP:conf/naacl/ShaLCS16,DBLP:conf/acl/HuangCFJVHS16,DBLP:conf/acl/Ahn17,DBLP:conf/cikm/0001RHZGHJL018}, and {\it event extraction}, which is to identify events with filled slots from a piece of news \cite{DBLP:conf/naacl/NguyenCG16,DBLP:conf/aaai/ShaQCS18,DBLP:conf/aaai/Liu00018,DBLP:conf/emnlp/0001Y00J18,DBLP:conf/acl/ChenXLZ015,DBLP:conf/acl/FengHTJQL16,DBLP:conf/emnlp/NguyenG16,DBLP:conf/emnlp/LiuLH18}.
Previous work focuses on extracting events from single news documents according to a set of pre-specified event types, such as arson, attack or earthquakes.

While useful for tracking highly specific types of events from news, the above setting can be relatively less useful for decision making in security and financial markets, which can require comprehensive knowledge on broad-coverage, fine-grained and dynamically-evolving event categories.
In addition, given the fact that different news agencies can report the same events, redundancy can be leveraged for better event extraction. 
In this paper, we investigate {\it open domain event extraction} (ODEE), which is to extract unconstraint types of events and induce universal event schemas from clusters of news reports.

\begin{figure}
	\centering
	\includegraphics[width=75mm]{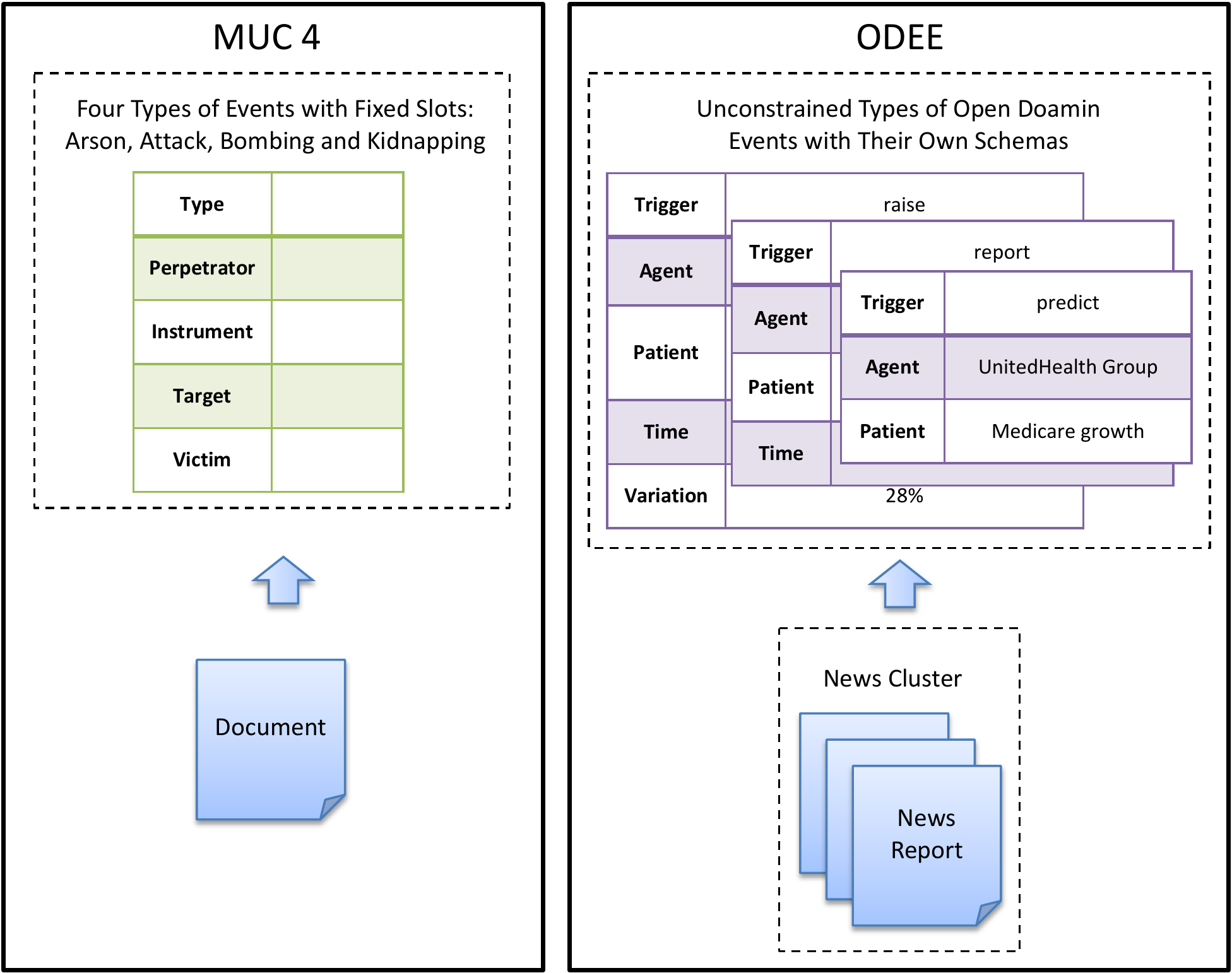}
	\caption{Comparison between MUC 4 and ODEE.}
	\label{fig:related_work}
\end{figure}

As shown in Figure \ref{fig:related_work}, compared with traditional event extraction task exemplified by MUC 4 \cite{DBLP:conf/muc/Sundheim92}, the task of ODEE poses additional challenges to modeling, which have not been considered in traditional methods.
First, more than one event can be extracted from a news cluster, where events can be flexible in having varying numbers of slots in the open domain, and slots can be flexible without identical distributions regardless of the event type, which has been assumed by previous work on schema induction.
Second, mentions of the same entities from different reports in a news cluster should be taken into account for improved performance.

We build an unsupervised generative model to address these challenges.
While previous work on generative schema induction \cite{DBLP:conf/emnlp/Chambers13,DBLP:conf/naacl/CheungPV13,DBLP:conf/acl/NguyenTFB15} relies on hand-crafted indicator features, we introduce latent variables produced by neural networks for better representation power.
A novel graph model is designed, with a latent event type vector for each news cluster from a global parameterized normal distribution, and textual redundancy features for entities.
Our model takes advantage of contextualized pre-trained language model (ELMo, \newcite{DBLP:conf/naacl/PetersNIGCLZ18}) and scalable neural variational inference \cite{srivastava17lda}.

To evaluate model performance, we collect and annotate a large-scale dataset from Google Business News\footnote{\url{https://news.google.com/?hl=en-US&gl=US&ceid=US:en}, crawled from Oct. 2018 to Jan. 2019.} with diverse event types and explainable event schemas.
In addition to the standard metrics for schema matching, we adapt {\it slot coherence} based on NPMI \cite{DBLP:conf/eacl/LauNB14} for quantitatively measuring the intrinsic qualities of slots and schemas, which are inherently clusters.

Results show that our neural latent variable model outperforms state-of-the-art event schema induction methods.
In addition, redundancy is highly useful for improving open domain event extraction. 
Visualizations of learned parameters show that our model can give reasonable latent event types.
To our knowledge, we are the first to use neural latent variable model for inducing event schemas and extracting events. 
We release our code and dataset at \url{https://github.com/lx865712528/ACL2019-ODEE}.


\section{Related Work}

The most popular schema induction and event extraction task setting is MUC 4, in which four event types - \textit{Arson}, \textit{Attack}, \textit{Bombing} and \textit{Kidnapping} - and four slots - \textit{Perpetrator}, \textit{Instrument}, \textit{Target} and \textit{Victim} - are defined.
We compare the task settings of MUC 4 and ODEE in Figure \ref{fig:related_work}.
For MUC 4, the inputs are single news documents, and the output belongs to four types of events with schemas consisting of fixed slots.
For ODEE, in contrast, the inputs are news clusters rather than the individual news, and the output is unconstrained types of open domain events and unique schemas with various slot combinations.

\textbf{Event Schema Induction} seminal work studies patterns \cite{DBLP:conf/naacl/ShinyamaS06,DBLP:conf/acl/FilatovaHM06,DBLP:conf/ijcnlp/QiuKC08} and event chains \cite{DBLP:conf/acl/ChambersJ11} for template induction.
For MUC 4, the current dominant methods include probabilistic generative methods \cite{DBLP:conf/emnlp/Chambers13,DBLP:conf/naacl/CheungPV13,DBLP:conf/acl/NguyenTFB15} that jointly model predicate and argument assignment, and ad-hoc clustering algorithms for inducing slots \cite{DBLP:conf/naacl/ShaLCS16,DBLP:conf/acl/HuangCFJVHS16,DBLP:conf/acl/Ahn17,DBLP:conf/cikm/0001RHZGHJL018}.
These methods all rely on hand-crafted discrete features without fully model the textual redundancy.
There are also works on modeling event schemas and scripts using neural language models \cite{DBLP:conf/conll/ModiT14,DBLP:conf/emnlp/RudingerRFD15,DBLP:conf/acl/PichottaM16}, but they do not explore neural latent variables and redundancy.

\textbf{Event Extraction} work typically assumes that event schemas are given, recognizing event triggers and their corresponding arguments.
This can be regarded as a subtask of ODEE.
Existing work exploits sentence-level \cite{DBLP:conf/acl/McCloskySM11,DBLP:conf/acl/LiJH13,DBLP:conf/acl/LiuCHL016,DBLP:conf/naacl/YangM16} and document-level statistics \cite{DBLP:conf/acl/LiaoG10,DBLP:conf/acl/JiG08,DBLP:conf/acl/HongZMYZZ11,DBLP:conf/naacl/ReichartB12}.
There has also been work using RNNs \cite{DBLP:conf/naacl/NguyenCG16,DBLP:conf/aaai/ShaQCS18,DBLP:conf/aaai/Liu00018,DBLP:conf/emnlp/0001Y00J18}, CNNs \cite{DBLP:conf/acl/ChenXLZ015,DBLP:conf/acl/FengHTJQL16,DBLP:conf/emnlp/NguyenG16} and GCNs \cite{DBLP:conf/emnlp/LiuLH18} to represent sentences of events.
Event extraction has been treated as a supervised or semi-supervised \cite{DBLP:conf/coling/LiaoG10,DBLP:conf/eacl/HuangR12} task.
In contrast, ODEE is a fully unsupervised setting.


\textbf{Event Discovery in Tweet Streams} extracts news-worthy clusters of words, segments and frames.
Both supervised and unsupervised methods have been used.
The former \cite{DBLP:conf/www/SakakiOM10,DBLP:conf/acl/BensonHB11} are typically designed to monitor certain event types, while the latter cluster features according to their burstiness \cite{DBLP:conf/icwsm/BeckerNG11,DBLP:conf/cikm/CuiZLMZ12,DBLP:conf/cikm/LiSD12,DBLP:conf/kdd/RitterMEC12,DBLP:conf/ijcnlp/QinZZZ13,DBLP:conf/www/IfrimSB14,DBLP:conf/clef/McMinnJ15,DBLP:conf/ialp/QinZZZ17}.
This line of work is similar to our work in using information redundancy, but different because we focus on formal news texts and induce structural event schemas.

\textbf{First Story Detection} (FSD) systems aim to identify news articles that discuss events not reported before.
Most work on FSD detects first stories by finding the nearest neighbors of new documents \cite{DBLP:conf/naacl/KumaranA05,DBLP:conf/sigir/MoranMMO16,DBLP:conf/coling/PanagiotouATKG16,DBLP:conf/sigir/VuurensV16}.
This line of work exploits textual redundancy in massive streams predicting whether or not a document contains a new event as a classification task.
In contrast, we study the event schemas and extract detailed events.

\section{Task and Data}



\noindent\textbf{Task Definition.}
In ODEE, the input consists of news clusters, each containing reports about the same event.
The output is a bag of open-domain events, each consisting of an event trigger and a list of event arguments in its own schema.
In most cases, one event is semantically sufficient to represent the output.

Formally, given an open-domain news corpus $\mathcal{N}$ containing a set of news clusters $\{ c \in \mathcal{N} \}$, suppose that there are $M_c$ news reports $\{ d_i \in c | i = 1, \dotsb, M_c \}$ in the news cluster $c$ focusing on the same event $ \mathcal{E}_c $.
The output is a pair ($ \mathcal{E}_c,\mathcal{T}_{\mathcal{E}}$), where $ \mathcal{E}_c $ is the aforementioned set of open-domain events and $\mathcal{T}_{\mathcal{E}}$ is a set of schemas that define the semantic slots for this set of events.

\noindent\textbf{Data Collection.}
We crawl news reports from Google Business News, which offers news clusters about the same events from different sources.
In each news cluster, there are no more than five news reports.
For each news report, we obtain the title, publish timestamp, download timestamp, source URL and full text.
In total, we obtain 55,618 business news reports with 13,047 news clusters in 288 batches from Oct. 17, 2018, to Jan. 22, 2019.
The crawler is executed about three times per day.
The full text corpus is released as \textit{GNBusiness-Full-Text}.
For this paper, we trim the news reports in each news cluster by keeping the title and first paragraph, releasing as \textit{GNBusiness-All}.

Inspired by the general slots in FrameNet \cite{DBLP:conf/acl/BakerFL98}, we design reference event schemas for open domain event types, which include eight possible slots: \textit{Agent}, \textit{Patient}, \textit{Time}, \textit{Place}, \textit{Aim}, \textit{Old Value}, \textit{New Value} and \textit{Variation}.
\textit{Agent} and \textit{Patient} are the semantic agent and patient of the trigger, respectively;
\textit{Aim} is the target or reason for the event.
If the event involves value changes, \textit{Old Value} serves the old value, \textit{New Value} serves the new value and \textit{Variation} is the variation between \textit{New Value} and \textit{Old Value}.
Note that the roles that we define are more thematic and less specific to detailed events as some of the existing event extraction datasets do \cite{DBLP:conf/muc/Sundheim92,DBLP:conf/lrec/NguyenTFB16}, because we want to make our dataset general and useful for a wide range of open domain conditions.
We leave finer-grained role typing to future work.

We randomly select 18 batches of news clusters, with 680 clusters in total, dividing them into a development set and a test set by a ratio of $1:5$.
The development set, test set and the rest unlabeled clusters are released as \textit{GNBusiness-Dev}, \textit{GNBusiness-Test} and \textit{GNBusiness-Unlabeled}, respectively.
One coauthor and an external annotator manually label the events in the news clusters as gold standards.
For each news cluster, they assign each entity which participants in the event or its head word a beforehand slot.
The inter-annotator agreement (IAA) for each slot realization in the development set has a Cohen's kappa \cite{cohen1960coefficient} $\kappa=0.7$.

\begin{table}
\centering
\small
\setlength{\tabcolsep}{1mm}{
\begin{tabular}{l|c|c|c|c}
\hline
\textbf{Split} & \textbf{\#C} & \textbf{\#R} & \textbf{\#S} & \textbf{\#W} \\
\hline
Test & 574 & 2,433 & 5,830 & 96,745 \\
Dev & 106 & 414 & 991 & 16,839 \\
Unlabelled & 12,305 & 52,464 & 127,416 & 2,101,558 \\
\hline
All & 12,985 & 55,311 & 134,237 & 2,215,142 \\
\hline
Full-Text & 12,985 & 55,311 & 1,450,336 & 31,103,698 \\
\hline
\end{tabular}
}
\vspace{-0.2cm}
\caption{\label{tab:statistics_data_splits} Data split statistics. ($C$ news clusters; $R$ news reports; $S$ sentences; $W$ words.)}
\end{table}

\begin{table}
\centering
\small
\begin{tabular}{l|c|c|c|c}
\hline
\textbf{Dataset} & \textbf{\#D} & \textbf{\#L} & \textbf{\#T} & \textbf{\#S} \\
\hline
MUC 4 & 1700 & 400 & 4 & 4 \\
ACE 2005 & 599 & 599 & 33 & 36 \\
ERE & 562 & 562 & 38 & 27 \\
ASTRE & 1038 & 100 & 12 & 18 \\
\textbf{GNBusiness} & 12,985 & 680 & -- & 8 \\
\hline
\end{tabular}
\vspace{-0.2cm}
\caption{\label{tab:dataset_statistics_comp} Comparison with existing datasets. ($D$ documents or news clusters; $L$ labeled documents or news clusters; $T$ event types; $S$ slots.)}
\end{table}

\begin{figure*}
\centering
\subfigure[\textit{ODEE-F}]{
\centering
\includegraphics[width=4cm]{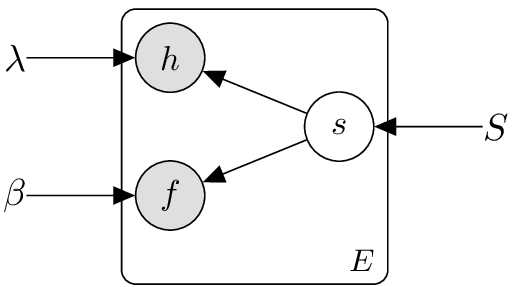}
\label{fig:odee-f}
}%
\subfigure[\textit{ODEE-FE}]{
\centering
\includegraphics[width=4cm]{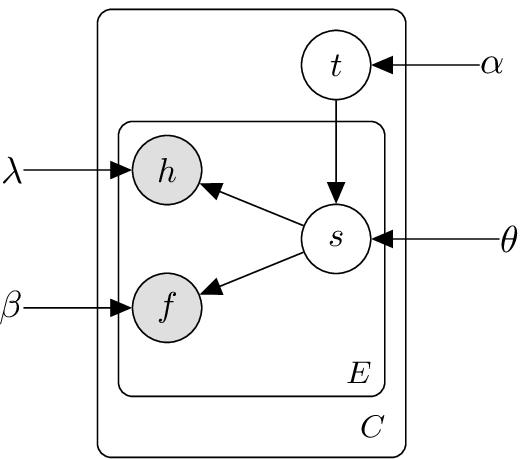}
\label{fig:odee-fe}
}%
\subfigure[\textit{ODEE-FER}]{
\centering
\includegraphics[width=4cm]{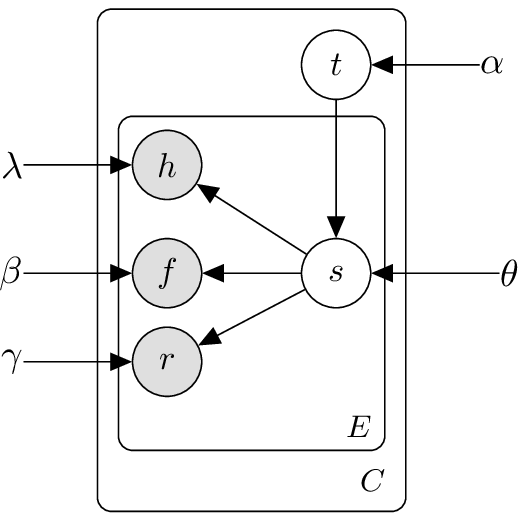}
\label{fig:odee-fer}
}
\caption{Plate notations for models.
($S$ -- \# of slots; $E$ -- \# of entities; $C$ -- \# of news clusters; $V$ -- head word vocabulary size; the grey circles are observed variables and the white circles are hidden variables.)}
\label{fig1}
\end{figure*}

The statistics of each data split is shown in Table \ref{tab:statistics_data_splits}, and a comparison with existing event extraction and event schema induction datasets, including ASTRE \cite{DBLP:conf/lrec/NguyenTFB16}, MUC 4, ACE 2005\footnote{\url{https://catalog.ldc.upenn.edu/LDC2006T06}} and ERE\footnote{\url{https://catalog.ldc.upenn.edu/LDC2013E64}}, is shown in Table \ref{tab:dataset_statistics_comp}.
Compared with the other datasets, GNBusiness has a much larger number of documents (i.e., news clusters in GNBusiness), and a comparable number of labeled documents.

\section{Method}

We investigate three incrementally more complex neural latent variable models for ODEE.

\subsection{Model 1}

Our first model is shown in Figure \ref{fig:odee-f}.
It can be regarded as a neural extension of \newcite{DBLP:conf/acl/NguyenTFB15}.
Given a corpus $\mathcal{N}$, we sample a slot $s$ for each entity $e$ from a uniform distribution of $S$ slots, and then a head word $h$ from a multinomial distribution, as well as a continuous feature vector $f \in \mathbb{R}^n$ produced by a contextual encoder.
For simplicity, we assume that $f$ follows a multi-variable normal distribution whose covariance matrix is a diagonal matrix.
We mark all the parameters (mean vectors and diagonal vectors of covariance matrixes) for the $S$ different normal distributions for $f$ as $\beta \in \mathbb{R}^{S \times 2n}$, where $n$ represents the dimension of $f$, treating the probability matrix $\lambda \in \mathbb{R}^{S \times V}$ in the slot-head distribution as parameters under the row-wise simplex constraint, where $V$ is the head word vocabulary size.
We call this model \textit{ODEE-F}.

\begin{algorithm}[t]
\caption{\textit{ODEE-F}}
\label{alg2}
\algsetup{indent=1em,linenosize=\footnotesize}
\footnotesize
\begin{algorithmic}[1]
	\FOR{each entity $e \in E$}
    	\STATE Sample a slot $s \sim \textrm{Uniform}(1, S)$
    	\STATE Sample a head $h \sim \textrm{Multinomial}(1, \lambda_s)$
    	\STATE Sample a feature vector $f \sim \textrm{Normal}(\beta)$
    \ENDFOR
\end{algorithmic}
\end{algorithm}

Pre-trained contextualized embeddings such as ELMo \cite{DBLP:conf/naacl/PetersNIGCLZ18}, GPTs \cite{radford2018improving,radford2019language} and BERT \cite{DBLP:journals/corr/abs-1810-04805} give improvements on a range of natural language processing tasks by offering rich language model information.
We choose ELMo\footnote{In practice, we use the ``small'' ELMo model with $2\times128$-d output in \url{https://allennlp.org/elmo} as initial parameters and fine-tune it on \textit{GNBusiness-Full-Text}.} as our contextual feature encoder, which manipulates unknown words by using character representations.

The generative story is shown in Algorithm \ref{alg2}.
The joint probability of an entity $e$ is
\begin{align*}
p_{\lambda,\beta}(e) = p(s) \times p_{\lambda}(h|s) \times p_{\beta}(f|s) \tag{1}
\end{align*}


\begin{algorithm}[t]
\caption{\textit{ODEE-FE}}
\label{alg3}
\algsetup{indent=1em,linenosize=\footnotesize}
\footnotesize
\begin{algorithmic}[1]
    \FOR{each news cluster $c \in \mathcal{N}$}
    	\STATE Sample a latent event type vector $t \sim \textrm{Normal}(\alpha)$
    	
    	\FOR{each entity $e \in E_c$}
    		\STATE Sample a slot $s \sim \textrm{Multinomial}(\textrm{MLP}(t;\theta))$
    		\STATE Sample a head $h \sim \textrm{Multinomial}(1, \lambda_s)$
    		\STATE Sample a feature vector $f \sim \textrm{Normal}(\beta_s)$
    	\ENDFOR
    \ENDFOR
\end{algorithmic}
\end{algorithm}

\subsection{Model 2}

A limitation of \textit{ODEE-F} is that sampling slot assignment $s$ from a global uniform distribution does not sufficiently model the fact that different events may have different slot distributions.
Thus, in Figure \ref{fig:odee-fe}, we further sample a latent event type vector $t \in \mathbb{R}^n$ for each news cluster from a global normal distribution parameterized by $\alpha$.
We then use $t$ and a multi-layer perceptron (MLP) with parameters $\theta$ to encode the corresponding slot distribution logits, sampling a discrete slot assignment $s \sim \textrm{Multinomial}(\textrm{MLP}(t;\theta))$.
The output of the MLP is passed through a softmax layer before being used.
We name this model as \textit{ODEE-FE}.

The generative story is shown in Algorithm \ref{alg3}.
The joint probability of a news cluster $c$ is
\begin{align*}
p_{\alpha,\beta,\theta,\lambda}(c) = & \,p_{\alpha}(t) \times \prod_{e \in E_c} p_{\theta}(s|t) \\
\times & \,p_{\lambda}(h|s) \times p_{\beta}(f|s) \tag{2}
\end{align*}

\subsection{Model 3}

Intuitively, the more frequently a coreferential entity shows up in a news cluster, the more likely it is with an important slot.
Beyond that, different news agencies focus on different aspects of event arguments, which can offer complementary information through textual redundancy.
One intuition is that occurrence frequency is a straightforward measure for word-level redundancy.
Thus, in Figure \ref{fig:odee-fer}, we additionally bring in the normalized occurrence frequency of a coreferential slot realization as an observed latent variable $r \sim \textrm{Normal}(\gamma_s)$.
We call this model \textit{ODEE-FER}.

\begin{algorithm}[t]
\caption{\textit{ODEE-FER}}
\label{alg4}
\algsetup{indent=1em,linenosize=\footnotesize}
\footnotesize
\begin{algorithmic}[1]
    \FOR{each news cluster $c \in \mathcal{N}$}
    	\STATE Sample a latent event type vector $t \sim \textrm{Normal}(\alpha)$
    	\FOR{each entity $e \in E_c$}
    		\STATE Sample a slot $s \sim \textrm{Multinomial}(\textrm{MLP}(t;\theta))$
    		\STATE Sample a head $h \sim \textrm{Multinomial}(1, \lambda_s)$
    		\STATE Sample a feature vector $f \sim \textrm{Normal}(\beta_s)$
    		\STATE Sample a redundancy ratio $r \sim \textrm{Normal}(\gamma_s)$
    	\ENDFOR
    \ENDFOR
\end{algorithmic}
\end{algorithm}

Formally, a news cluster $c$ receives a latent event type vector $t$ where each entity $e \in E_c$ receives a slot type $s$.
The generative story is shown in Algorithm \ref{alg4}.
The joint distribution of a news cluster with head words, redundant contextual features and latent event type is
\begin{align*}
p_{\alpha,\beta,\gamma,\theta,\lambda}(c) = & \,p_{\alpha}(t) \times \prod_{e \in E_c} p_{\theta}(s|t) \\
\times & \,p_{\lambda}(h|s) \times p_{\beta}(f|s) \times p_{\gamma}(r|s) \tag{3}
\end{align*}

\subsection{Inference}

We now consider two tasks for \textit{ODEE-FER}: (1) learning the parameters 
and (2) performing inference to obtain the posterior distribution of the latent variables $s$ and $t$, given a news cluster $c$.
We adapt the amortized variational inference method of \newcite{srivastava17lda}, using neural inference network to learn the variational parameters.
For simplicity, we concatenate $f$ with $r$ as a new observed feature vector $f'$ in \textit{ODEE-FER} and merge their parameters as $\beta' \in \mathbb{R}^{S \times (2n+2)}$.

\begin{figure}
	\centering
	\includegraphics[width=65mm]{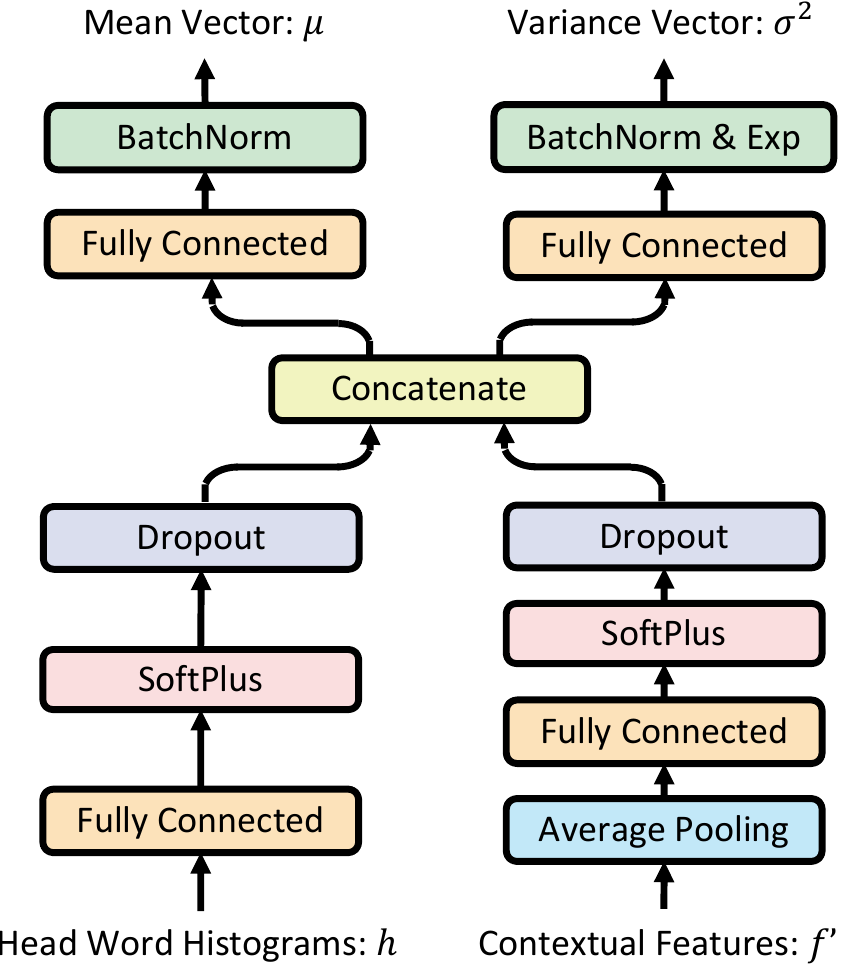}
	\caption{The framework of our inference network.}
	\label{fig:inference_network}
\end{figure}

Following \newcite{srivastava17lda}, we collapse the discrete latent variable $s$ to obtain an Evidence Lower BOund (ELBO) \cite{kingma2013auto} of the log marginal likelihood:
\begin{align*}
&\textrm{log}\,p_{\alpha,\beta',\theta,\lambda}(c)\\
&=\textrm{log}\,\int_{t}[\prod_{e \in E_c} \, p_{\lambda,\theta}(h|t) \, p_{\beta',\theta}(f'|t)] \, p_{\alpha}(t) \, dt \\
&\geq \textrm{ELBO}_{c}(\alpha,\beta',\theta,\lambda,\omega)\\
&=\mathbb{E}_{q_{\omega}(t)}\textrm{log}\,p_{\beta',\theta,\lambda}(c|t) - D_{\textrm{KL}}[ q_{\omega}(t) \| p_{\alpha}(t) ] \tag{4}\label{eq13}
\end{align*}
where $D_{\textrm{KL}}[q_{\omega} \| p_{\alpha}]$ is the KL divergence between the variational posterior $q_{\omega}$ and the prior $p_{\alpha}$.
Due to the difficulty in computing the KL divergence between different categories of distributions and the existence of simple and effective reparameterization tricks for normal distributions, we choose $q_{\omega}(t)$ to be a normal distribution parameterized by $\omega$, which is learned by a neural inference network.
As shown in Figure \ref{fig:inference_network}, our inference network takes the head word histograms $h$ (the times of each head word appears in a news cluster) and contextual features $f'$ as inputs, and computes the mean vector $\mu$ and the variance vector $\sigma^2$ of $q_{\omega}(t)$.

Equation \ref{eq13} can be solved by obtaining a Monte Carlo sample and applying reparameterization tricks for the first term, and using the closed-form for the KL divergence term.
We then use the ADAM optimizer \cite{DBLP:journals/corr/KingmaB14} to maximumize the ELBO.
In addition, to alleviate the component collapsing problem \cite{dinh2016training},
we follow \newcite{srivastava17lda} and use high moment weight ($>0.8$) and learning rate (in $[0.001, 0.1]$) in the ADAM optimizer, performing batch normalization \cite{DBLP:conf/icml/IoffeS15} and dropout \cite{DBLP:journals/jmlr/SrivastavaHKSS14}.
After learning the model, we make slot assignment for each entity mention by MLE, choosing the slot $s$ that maximizes the likelihood
\begin{align*}
p_{\beta',\theta,\lambda}(s|e,t) \propto& \,p_{\beta',\theta,\lambda}(s,h,f',t) \\
=& \,p_{\theta}(s|t) \times p_{\lambda}(h|s) \times p_{\beta'}(f'|s) \tag{5}\label{eq14}
\end{align*}

\subsection{Assembling Events for Output}

To assemble the events in a news cluster $c$ for final output, we need to find the predicate for each entity, which now has a slot value.
We use POS-tags and parse trees produced by the Stanford dependency parser \cite{DBLP:conf/acl/KleinM03} to extract the predicate for the head word of each entity mention.
The following rules are applied:
(1) if the governor of a head word is \textit{VB}, or
(2) if the governor of a head word is \textit{NN} and belongs to the \textit{noun.ACT} or \textit{noun.EVENT} category of WordNet, then it is regarded as a predicate.

We merge the predicates of entity mentions in the same coreference chain as a predicate set.
For each predicate $v$ in these sets, we find the entities whose predicate set contains $v$, treating the entities as arguments of the event triggered by $v$.
Finally, by ranking the numbers of arguments, we obtain top-N open-domain events as the output $\mathcal{E}_c$.

\section{Experiments}

We verify the effectiveness of neural latent variable modeling and redundancy information for ODEE, and conduct case analysis.
All our experiments are conducted on the GNBusiness dataset.
Note that we do not compare our models and existing work on MUC 4 or ACE 2005 due to the fact that these datasets do not consist of news clusters.




\vspace{1mm}
\noindent\textbf{Settings.}
The hyper-parameters in our models and inference network are shown in Table \ref{tab:setting}.
Most of the hyper-parameters directly follow \newcite{srivastava17lda}, while the slot number $S$ is chosen according to development experiments.

\begin{table}
\centering
\small
\begin{tabular}{c|c}
\hline
\textbf{Name} & \textbf{Value} \\
\hline
Slots number $S$ & 30 \\
Feature Dimension $n$ & 256 \\
Fully connected layer size & 100 \\
MLP layer number & 1 \\
Activation function & softplus \\
Learning rate & 0.002 \\
Momentum & 0.99 \\
Dropout rate & 0.2 \\
Batch size & 200 \\
\hline
\end{tabular}
\vspace{-0.1cm}
\caption{\label{tab:setting} Hyper-parameters setting.}
\end{table}

\subsection{Evaluation Metrics}

\noindent\textbf{Schemas Matching.}
We follow previous work and use \textit{precision}, \textit{recall} and \textit{F1-score} as the metrics for schema matching \cite{DBLP:conf/acl/ChambersJ11,DBLP:conf/emnlp/Chambers13,DBLP:conf/naacl/CheungPV13,DBLP:conf/acl/NguyenTFB15,DBLP:conf/naacl/ShaLCS16,DBLP:conf/acl/Ahn17}.
The matching between model answers and references is based on the head word.
Following previous work, we regard as the head word the right-most word of an entity phrase or the right-most word before the first ``of'', ``that'', ``which'' and ``by'' if any.

In addition, we also perform slot mapping, between slots that our model learns and slots in the annotation.
Following previous work on MUC 4 \cite{DBLP:conf/emnlp/Chambers13,DBLP:conf/naacl/CheungPV13,DBLP:conf/acl/NguyenTFB15,DBLP:conf/naacl/ShaLCS16,DBLP:conf/acl/Ahn17}, we implement automatic greedy slot mapping.
Each reference slot is mapped to a learned slot that ranks the best according to the \textit{F1-score} metric on \textit{GNBusiness-Dev}.

\vspace{1mm}
\noindent\textbf{Slot Coherence.}
Several metrics of qualitative topic coherence evaluation have been proposed.
\newcite{DBLP:conf/eacl/LauNB14} showed that normalized point-wise mutual information (NPMI) between all the pairs of words in a set of topics the most closely matches human judgment among all the competing metrics.
We thus adopt it as \textit{slot coherence}\footnote{We use the implementation in \url{https://github.com/jhlau/topic_interpretability}.}.

Formally, the slot coherence $C_{\textrm{NPMI}}(s)$ of a slot $s$ is calculated by using its top-N head words as
\begin{align*}
&C_{\textrm{NPMI}}(s)=\frac{2}{N^2-N}\sum_{i=2}^{N}\sum_{j=1}^{i-1}\textrm{NPMI}(w_i, w_j) \tag{6}\label{eq17} \\
&\textrm{NPMI}(w_i, w_j)=\frac{\textrm{log}\frac{p(w_i, w_j)+\epsilon}{p(w_i)\cdot p(w_j)}}{-\textrm{log}(p(w_i, w_j)+\epsilon)} \tag{7}\label{eq18}
\end{align*}
where $p(w_j)$ and $p(w_i, w_j)$ are estimated based on word co-occurrence counts derived within a sliding window over external reference documents and $\epsilon$ is added to avoid zero logarithm.

Previous work on topic coherence uses Wikipedia and Gigaword as the reference corpus to calculate word frequencies \cite{DBLP:conf/naacl/NewmanLGB10,DBLP:conf/eacl/LauNB14}.
We use \textit{GNBusiness-Full-Text}, in which there are 1.45M sentences and 31M words, which is sufficient for estimating the probabilities.
To reduce sparsity, for each news report, we count word co-occurrences in the whole document instead of a sliding window.
In addition, for each slot, we keep the top-5, top-10, top-20, and top-100 head words, averaging the $4 \times S$ coherence results over a test set.

\begin{figure}
	\centering
	\includegraphics[width=65mm]{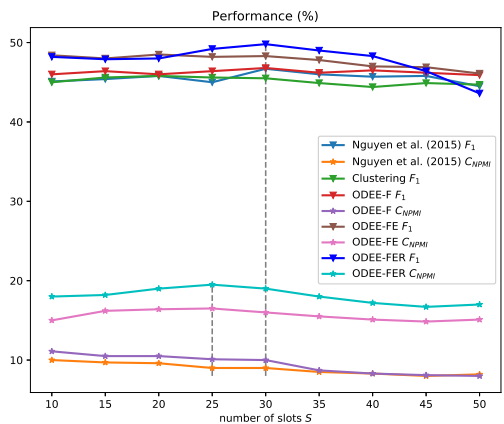}
	\caption{$F_1$ scores of schemas matching and averaged slot coherences $C_{\textrm{NPMI}}$ of the five models with different numbers of slots $S$.
	}
	\label{fig:hyps}
\end{figure}

\subsection{Development Experiments}

We learn the models on \textit{GNBusiness-All} and use \textit{GNBusiness-Dev} to determine the slot number $S$ by grid search in $[10, 50]$ with the step equals to $5$.
Figure \ref{fig:hyps} shows the $F_1$ scores of schemas matching and averaged slot coherences of the five models we introduce in the next subsection with different numbers of slots $S$ ranging from $10$ to $50$.
We can see that for the best $F_1$ score of \textit{ODEE-FER}, the optimal number of slots is 30, while for the best slot coherence, the optimal number of slots is 25.
A value of $S$ larger than 30 or smaller than 25 gives lower results on both $F_1$ score and slot coherence.
Considering the balance between $F_1$ score and slot coherence, we chose $S=30$ as our final $S$ value for the remaining experiments.

\subsection{Final Results}

\begin{table}
\centering
\small
\begin{tabular}{l|ccc}
\hline
\textbf{Method} & \multicolumn{3}{c}{\textbf{Schema Matching (\%)}} \\
 & $P$ & $R$ & $F_1$ \\
\hline
\newcite{DBLP:conf/acl/NguyenTFB15} & 41.5 & 53.4 & 46.7 \\
Clustering & 41.2 & 50.6 & 45.4 \\
\hline
ODEE-F & 41.7 & 53.2 & 46.8 \\
ODEE-FE & 42.4 & 56.1 & 48.3 \\
ODEE-FER & \textbf{43.4} & \textbf{58.3} & \textbf{49.8} \\
\hline
\end{tabular}
\vspace{-0.1cm}
\caption{\label{tab:slot_matching} Overall performance of schema matching.}
\end{table}

\begin{table}
\centering
\small
\begin{tabular}{l|c}
\hline
\textbf{Method} & \textbf{Ave Slot Coherence} \\
\hline
\newcite{DBLP:conf/acl/NguyenTFB15} & 0.10 \\
\hline
ODEE-F & 0.10 \\
ODEE-FE & 0.16 \\
ODEE-FER & \textbf{0.18} \\
\hline
\end{tabular}
\vspace{-0.1cm}
\caption{\label{tab:slot_coherence} Averaged slot coherence results.}
\end{table}

Table \ref{tab:slot_matching} and Table \ref{tab:slot_coherence} show the final results.
The $p$ values based on the appropriate t-test are provided below in cases where the compared values are close.
We compare our work with \newcite{DBLP:conf/acl/NguyenTFB15}, the state-of-the-art model on MUC 4 representing each entity as a triple containing a head word, a list of attribute relation features and a list of predicate relation features.
Features in the model are discrete and extracted from dependency parse trees.
The model structure is identical to our \textit{ODEE-F} except for the features.

To test the strengths of our external features in isolation, we build another baseline model by taking the continuous features of each entity in \textit{ODEE-F} and runing spectral clustering \cite{DBLP:journals/sac/Luxburg07}.
We call it \textit{Clustering}.

\vspace{1mm}
\noindent\textbf{Schemas Matching.}
Table \ref{tab:slot_matching} shows the overall performance of schema matching on \textit{GNBusiness-Test}.
From the table, we can see that \textit{ODEE-FER} achieves the best $F_1$ scores among all the methods.
By comparing \textit{\newcite{DBLP:conf/acl/NguyenTFB15}} and \textit{ODEE-F} ($p=0.01$), we can see that using continuous contextual features gives better performance than discrete features.
This demonstrates the advantages of continuous contextual features for alleviating the sparsity of discrete features in texts.
We can also see from the result of \textit{Clustering} that using only the contextual features is not sufficient for ODEE, while combining with our neural latent variable model in \textit{ODEE-F} can achieve strong results ($p=6 \times 10^{-6}$).
This shows that the neural latent variable model can better explain the observed data.

These results demonstrate the effectivenesses of our method in incorporating with contextual features, latent event types and redundancy information.
Among ODEE models, \textit{ODEE-FE} gives a $2\%$ gain in $F_1$ score against \textit{ODEE-F}, which shows that the latent event type modeling is beneficial and the slot distribution relies on the latent event type.
Additionally, there is a $1\%$ gain in $F_1$ score by comparing \textit{ODEE-FER} and \textit{ODEE-FE} ($p=2 \times 10^{-6}$), which confirms that leveraging redundancy is also beneficial in exploring which slot an entity should be assigned.

\vspace{1mm}
\noindent\textbf{Slot Coherence.}
Table \ref{tab:slot_coherence} shows the comparison of averaged slot coherence results over all the slots in the schemas.
Note that we do not report the slot coherence for the \textit{Clustering} model because it does not output the top-N head words in each slot.
The averaged slot coherence of \textit{ODEE-FER} is the highest, which is consistent with the conclusion from Table \ref{tab:slot_matching}.
The averaged slot coherence of \textit{ODEE-F} is comparable to that of \textit{\newcite{DBLP:conf/acl/NguyenTFB15}} ($p=0.3415$), which again demonstrates that the contextual features are a strong alternative to discrete features.
The scores of \textit{ODEE-FE} ($p=0.06$) and \textit{ODEE-FER} ($p=10^{-5}$) are both higher than that of \textit{ODEE-F}, which proves that the latent event type is critical in ODEE.

\subsection{Latent Event Type Analysis}

\begin{figure}
	\centering
	\includegraphics[width=75mm]{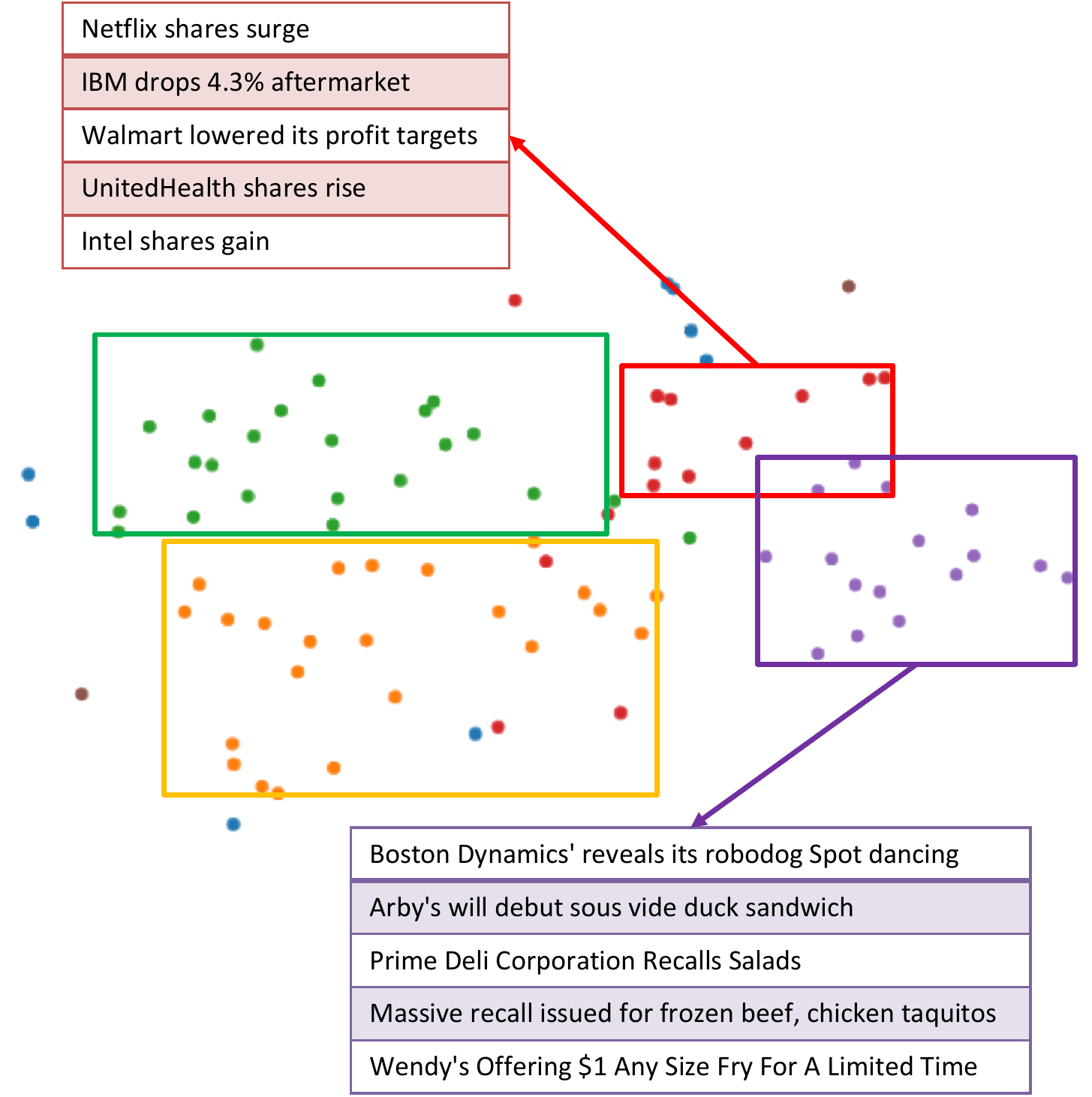}
	\caption{T-SNE visualization results of the latent event type vectors in the test set with colored labels produced by spectral clustering.}
	\label{fig:t-SNE}
\end{figure}


We are interested in learning how well the latent event type vectors can be modeled.
To this end, for each news cluster in \textit{GNBusiness-Dev}, we use our inference network in Figure \ref{fig:inference_network} to calculate the mean $\mu$ for the latent event type vector $t$.
T-SNE transformation \cite{maaten2008visualizing} of the mean vectors are shown in Figure \ref{fig:t-SNE}.
Spectral clustering is further applied, and the number of clusters is chosen by the Calinski-Harabasz Score \cite{calinski1974dendrite} in grid search. 

In Figure \ref{fig:t-SNE}, there are four main clusters marked in different colors.
Representative titles of news reports are shown as examples.
We find that the vectors show salient themes for each main cluster.
For example, the red cluster contains news reports about rise and drop of stocks such as \textit{Netflix shares surge}, \textit{IBM drops}, \textit{Intel shares gain}, etc;
the news reports in the purple cluster are mostly about product related activities, such as \textit{Boston Dynamics' reveals its robodog Spot dancing}, \textit{Arby's will debut sous vide duck sandwich}, \textit{Wendy's Offering \$1 Any Size Fry}, etc.
The green cluster and the orange cluster are also interpretable.
The former is about organization reporting changes, while the latter is about service related activities.


\subsection{Case Study}

\begin{figure}
	\centering
	\includegraphics[width=75mm]{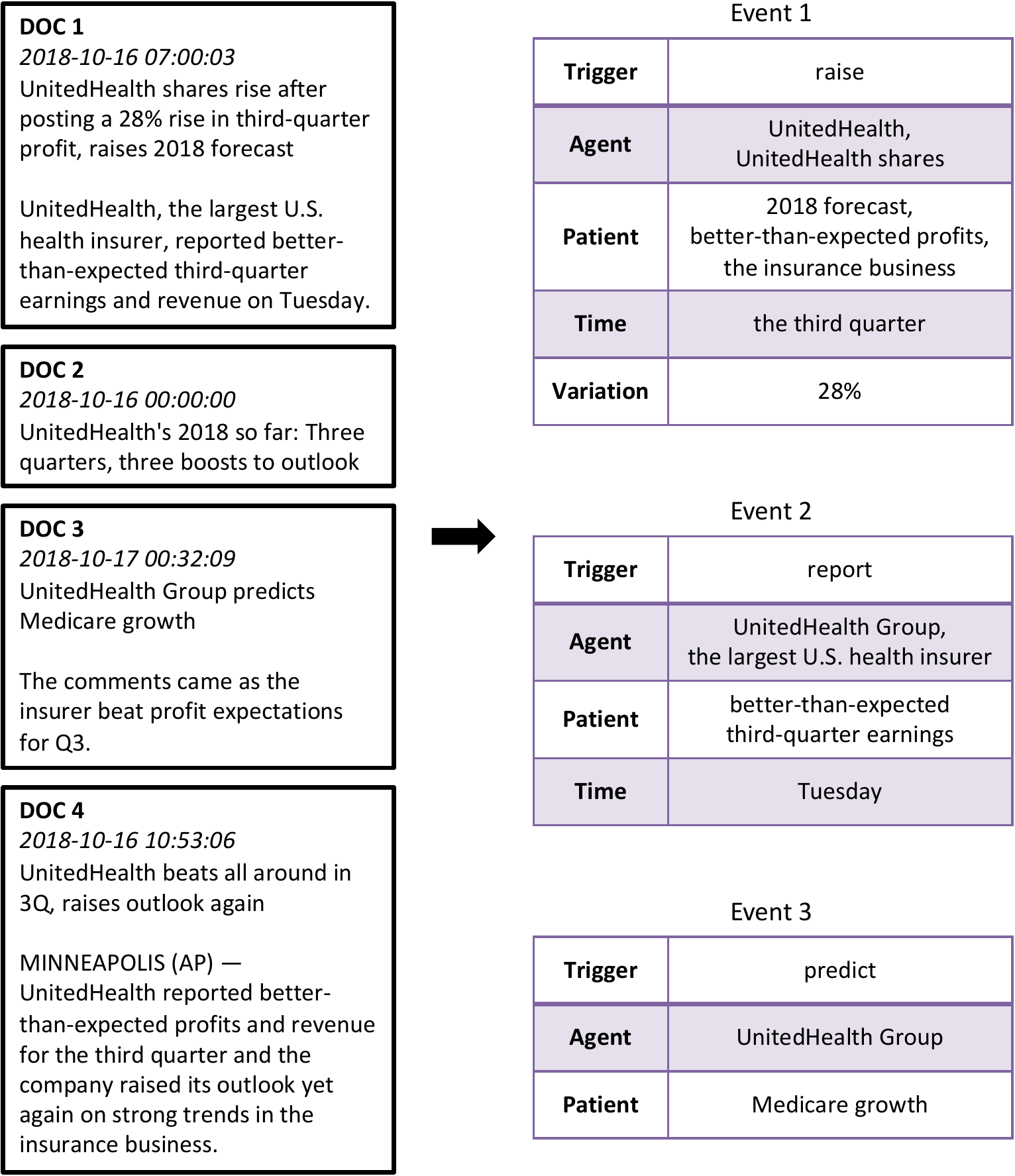}
	\caption{Extracted open domain events for \textit{UnitedHealth shares rise}.}
	\label{fig:case_study}
\end{figure}

We further use the news cluster \textit{UnitedHealth shares rise} in Figure \ref{fig:t-SNE} for case study.
Figure \ref{fig:case_study} shows the top-3 open-domain events extracted from the news cluster, where four input news reports are shown on the left and three system-generated events are shown on the right with mapped slots.

By comparing the plain news reports and the extracted events, we can see that the output events give a reasonable summary for the news cluster with three events triggered by ``raise'', ``report'' and ``predict'', respectively.
Most of the slots are meaningful and closely related to the trigger, while covering most key aspects.
However, this example also contains several incorrect slots.
In the event 1, the slot ``Variation'' and its realization ``28\%'' are only related to the entity ``better-than-expected profits'', but there are three slot realizations in the event, which causes confusion.
In addition, the slot ``Aim'' does not appear in the first event, whose realization should be ``third-quarter profit'' in document 1.
The reason may be that we assemble an event only using entities with the same predicate, which introduces noise.
Besides, due to the preprocessing errors in resolving coreference chains, some entity mentions are missing from the output.

There are also cases where one slot realization is semantically related to one trigger but eventually appears in a different event.
One example is the entity ``better-than-expected profits'', which is related to the predicate word ``report'' but finally appears in the ``raise'' event.
The cause can be errors propagated from parsing dependency trees, which confuse the syntactic predicate of the head word of an entity.

\section{Conclusion}

We presented the task of open domain event extraction, extracting unconstraint types of events from news clusters.
A novel latent variable neural model was investigated, which explores latent event type vectors and entity mention redundancy.
In addition, GNBusiness dataset, a large-scale dataset annotated with diverse event types and explainable event schemas, is released along with this paper.
To our knowledge, we are the first to use neural latent variable model for inducing event schemas and extracting events.

\section*{Acknowledgments}
We thank the anonymous reviewers for their valuable comments and suggestions.
We thank Kiem-Hieu Nguyen from Hanoi University of Science and Technology for providing source code and solving confusions for their work.
We thank Katherine Keith from University of Massachusetts at Amherst for sharing valuable experiences on probabilistic models.
This work is supported by National Natural Science Foundation of China No. 61751201, National Key Research and Development Plan No. 2016QY03D0602, Research Foundation of Beijing Municipal Science and Technology Commission No. Z181100008918002, the funding from Rxhui Inc\footnote{\url{https://rxhui.com}} and China Scholarship Council No. 201806030142.
The work is done when Xiao Liu is visiting Yue Zhang.

\bibliography{acl2019}
\bibliographystyle{acl_natbib}

\end{document}